\documentclass[runningheads]{llncs}

\usepackage{mathtools}
\usepackage{amsmath}
\usepackage{amssymb}
\usepackage{multirow}
\usepackage{graphicx}
\usepackage{epstopdf}
\usepackage{tabularx}
\usepackage{subcaption}

\usepackage{svg}

\usepackage{xcolor}

\usepackage{enumitem}
\usepackage{url}

\usepackage{cleveref}
\Crefname{algorithm}{Alg.}{Algs.}
\Crefname{equation}{Eq.}{Eqs.}
\Crefname{section}{Sec.}{Secs.}
\Crefname{figure}{Fig.}{Figs.}
\Crefname{tabular}{Tab.}{Tabs.}
\Crefname{table}{Tab.}{Tabs.}

\usepackage{algorithm}
\usepackage{algorithmic}

\def\MYCALL#1#2{{\small\textsc{#1}}(\textup{#2})}

\def\MYNIL{\textsc{Nil}}

\begin{document}
\title{Scope Compliance Uncertainty Estimate}
%
%
\author{Al-Harith Farhad\inst{1}\orcidID{0000-0001-6749-0620 } \and
Ioannis Sorokos\inst{2}\orcidID{0000-0003-2704-8381} \and
Mohammed Naveed Akram\inst{2}\orcidID{0000-0002-0924-5536} \and 
Koorosh Aslansefat \inst{3}\orcidID{0000-0001-9318-8177} \and
Daniel Schneider\inst{2}\orcidID{0000-0003-3465-9738}
}
\authorrunning{H. Farhad et al.}
%
\institute{University of Mannheim, Schloss, 68131 Mannheim, Germany \email{afarhad@mail.uni-mannheim.de} \and
Fraunhofer IESE, Fraunhofer-Platz 1, 67663 Kaiserslautern, Germany \email{\{ioannis.sorokos, andreas.schmidt, naveed.akram, daniel.schneider\}@iese.fraunhofer.de} \and
University of Hull, Cottingham Rd., HU6 7RX Hull, UK \email{k.aslansefat@hull.ac.uk}
}
\maketitle              
\begin{abstract}

The zeitgeist of the digital era has been dominated by an expanding integration of Artificial Intelligence~(AI) in a plethora of applications across various domains. With this expansion, however, questions of the safety and reliability of these methods come have become more relevant than ever.
SafeML is a model-agnostic approach for performing such monitoring, using distance measures based on statistical testing of the training and operational datasets; comparing them to a predetermined threshold, returning a binary value whether the model should be trusted in the context of the observed data or be deemed unreliable.
Although a systematic framework exists for this approach, its performance is hindered by: (1) a dependency on a number of design parameters that directly affect the selection of a safety threshold and therefore likely affect its robustness, (2) an inherent assumption of certain distributions for the training and operational sets, as well as (3) a high computational complexity for relatively large sets. 
This work addresses these limitations by changing the binary decision to a continuous metric. Furthermore, all data distribution assumptions are made obsolete by implementing non-parametric approaches, and the computational speed increased by introducing a new distance measure based on the Empirical Characteristics Functions~(ECF).

\keywords{Machine Learning \and Monitoring \and Safety \and Uncertainty}
\end{abstract}
\section{Introduction}
\label{sec:intro}
The last decade has seen a progression of paradigm-shifting developments in long-standing problems in AI. These developments have allowed AI methods to be used in areas as diverse as computer vision, natural language processing, and other. Adding to this progression, concerns of the safety of such systems are also raised, as AI is integrated into safety-critical systems such as autonomous vehicles, tumor detection, etc. Therefore, an increasing need for suitable safety assurance approaches quickly emerges, especially given that recent reviews show that both the frequency and the severity of future AI failures will steadily increase \cite{aiFailure}.

One of the facets of machine learning~(ML) robustness is the estimation of the confidence affiliated with the output of a given ML model. Even though many ML and AI are equipped with inherent heuristics of uncertainty, or a probabilistic output that could be interpreted as a level of confidence. Such an output alone is not sufficient to confirm and substantiate the trustworthiness of the model, as it would be trained along with the model using the same data, making it prone to the same noise and biases.


A safety argument for a system with ML components is expected to be tailored to a specific application and its context and comprise of a diverse range of measures and assumptions \cite{afarhad2022safeml}. Some of these requirements include development-time approaches and runtime approaches. SafeML, proposed in \cite{Aslansefat2020safeml} and improved in \cite{aslansefat2021toward,afarhad2022safeml}, is an approach that can address the safety of machine learning models at runtime by ensuring they work in the intended context. This is achieved by comparing training and operational data of the machine learning model in question and determining whether they are statistically too dissimilar to yield a trustworthy answer.

This paper demonstrates problems and limitations with the current implementation of statistical distance methods in SafeML, and proposes approaches to resolve these issues. 
\subsection{SafeML}
SafeML is a collection of statistical distance measures that estimate the distance between the training (used to train the ML model) and operational datasets (provided as input to the model at runtime), using the \emph{Empirical Cumulative Distribution Function}~(ECDF) \cite{afarhad2022safeml}. While similar in principle, each of the statistical distance measures focus on a certain aspect of the ECDF and thus would perform differently depending on the given distribution. Hence, each of the distance measures can be thought of as a metric of dissimilarity between the ECDFs \cite{afarhad2022safeml,akram_stadre_2022} and will be referred to as Statistical Distance Dissimilarity~(SDD).
To increase the confidence in the validity of the statistical measures, a bootstrap-based p-value evaluation method is added to the measurements, as in~\cite{aslansefat2021toward}.

The estimated SDD has been shown to negatively correlate with the accuracy of the underlying ML model in \cite{Aslansefat2020safeml}. The authors also proposed a workflow in the same paper to apply SafeML for the monitoring of ML models. The task is divided into two phases, an offline/training phase and an online/application phase.
A model is trained on a trusted dataset in the training phase. Then, that model is deployed and a sufficient number of samples from inputs are buffered until they are enough to estimate a reliable statistical distance between the ECDFs. If the outcome significantly exceeds an acceptable threshold value, alternative actions can be taken, e.g. operator intervention. The original workflow in \cite{Aslansefat2020safeml} did not provide guidance on the choice of the number of samples needed for a trustworthy prediction, nor the choice of the acceptance threshold.
A framework was proposed in \cite{afarhad2022safeml} to systematically estimate the number of samples needed and the acceptance threshold.

In short, SafeML only relies on the comparison between training and operational data of the ML model in question and it determines whether they are statistically too dissimilar to yield a trustworthy prediction. Therefore, it can be deployed independently of the application domain. As such, it was used for security attack detection in the original paper using the dataset \cite{sharafaldin2018toward}, in time series data for stock price prediction \cite{akram_stadre_2022} and finally in this paper for image classification using the \emph{German Traffic Sign Recognition Benchmark}~(GTSRB) \cite{stallkamp2012man}.

\subsection{Motivation}

As with any field of application, no silver bullet or best approach to solve the issue of ML model safety exists. Each method can be thought of as a step or a building block towards achieving that goal. To this end, ensuring the robustness of SafeML by investigating its limitations and allowing its methods to be compatible with other existing methods, such as uncertainty wrappers \cite{klas2019uncertainty}, is one of the steps to achieve the goal of model safety.

\subsection{Contribution and Outline}

The contribution of the paper is three-fold.
First, we use a bootstrapped approach for power analysis to eliminate any assumptions about the underlying distribution. It will be validated by comparing it to the result of a parametric test, used in \cite{afarhad2022safeml}, on a synthetic dataset with a known parameters.
Secondly, we added a new distance measure, the Epps and Singleton Test, which relies on the empirical characteristics function rather than the empirical cumulative distribution function.
Finally, the binary output of SafeML is extended into a bounded uncertainty estimate for the likelihood of out-of-distribution samples, and later evaluated on the German Traffic Sign Recognition Benchmark.

The remainder of the paper is as follows: In \Cref{sec:background}, we discuss background and related work, including approaches both similar, and differing to SafeML.
In \Cref{sec:methodology}, we describe our approach for extending the binary output of SafeML into a continuous metric, applying a bootstrap method for power analysis.
In \Cref{sec:experiment}, we discuss our experiment setup for evaluation and the results obtained before we recap our key points and discuss future work in \Cref{sec:conclusion}.

\section{Background and Related Work}
\label{sec:background}

While AI safety is relatively new, and a hot topic in both practice and academia, a lot of efforts have already been made to advance in it. According to \cite{shafaei_uncertainty_2018}, the uncertainty relating to the safety of a ML system can be classified as one of 2 types: 
\begin{itemize}
    \item \textbf{aleatoric uncertainty:} relating to data-dependent issues in ML such as the noise inherent in the training dataset

    \item \textbf{epistemic uncertainty:} relating to model-dependent issues such as the ambiguity of the model's output in operation
\end{itemize}

Reviewing the state-of-the-art literature in safety-critical systems that rely on machine learning, as well as safety of AI applications in general, reveals two main types of safety assurance, development-time safety based on model checking and verification, and runtime safety based on uncertainty estimation and confidence guarantees \cite{henne2020benchmarking}. The methods discussed here are not exhaustive, but are a selection of the methods in the fields of AI safety that share similarities in theory or implementation with our approach. However there could exist other hybrid methods that provide both development-time and runtime safety.

\subsection{Development-time ML Safety}
\label{sec:devtime}
As SafeML is a runtime method, it is inherently distinct from all the approaches mentioned in this section. But unlike SafeML, which does not provides no guarantees for the behavior of the model before deployment, these approaches do attempt to certify and verify certain aspects of the model.

Aiming to certify the robustness of neural networks, a novel abstract domain was introduced in \cite{singh2019certifyNN}. Before model deployment, the functions within the network, such as ReLU, sigmoid, tanh, and maxpool activations, would be transformed into the novel domain and the input data are tested, and certified against different transformations. Compared to SafeML, this approach is highly specific as only certain activations can be transformed

%
In the same vein, Markov Decision Process (MDP)  were proposed to be paired with neural networks to verify their robustness through statistical model checking \cite{Gros2020Modelchecking}, where a MDP is used as a context to analyse an externally learned neural network is used as a determinizer as the choices of the MDP are otherwise non-deterministic. DeepImportance, proposed in \cite{gerasimou2020importance}, is also a model-specific solution that presents new Importance-Driven Criteria as layer-wise functions to be evaluated during the testing procedure, thereby providing a systematic framework for ML testing. However, this approach is model specific as well.

The final more common approach to secure machine learning models is the exploration and investigation of every possible input perturbation, to produce safe and robust solutions and verification of ML models \cite{ruoss2020efficient,fischer2019dl2,mirman2019provable}. In comparison to SafeML, these proofs are all model specific.

While all of the methods mentioned above can be seen as a type of verification of certain aspects of ML components, a formal verification of most machine learning models is not possible,
 as the specification of low-level components cannot be feasibly applied due to very high number of possible inputs, parameters, and variable environment.

\subsection{Runtime ML Safety}
\label{sec:runsafe}

Most methods that assure safety at runtime try to estimate the uncertainty of the prediction of the model, providing a metric or an indicator 
that the output is correct. One of those metrics is a commonality metric proposed in \cite{paterson2021detection}, which inspects the second-to-last layer of a {Deep Neural Network}~(DNN). This commonality metric is calculated from the ratio of the activation outputs of the neurons in the last layer during training (across all training instances), versus their activation during operation, for the given operational input. While this approach have similar ideas in common with SafeML, it diverges in that it is model-specific, as the metric uses the output of the last layer, and relies on the assumption that the activation pattern and classification are correlated. 
Instead, SafeML is completely independent from the underlying model, and makes no assumption on the distribution of the training and operational data. 

Another estimator is the approach proposed in \cite{klas2020framework,klas2019uncertainty}. These uncertainty wrappers divide the uncertainty in machine learning into 3 separate components, namely (1) model performance, (2) input quality, and (3) scope compliance, providing a set of useful functions to evaluate the existing uncertainties in each of the separate components. When compared to SafeML, the third layer, scope compliance, i.e both of them are model-agnostic. 

Finally, the NN-dependability-kit \cite{cheng2019nn} proposed a novel dependability metric along with a formal reasoning engine for ensuring that the generalization of neural networks does not lead to undesired behaviors. These are then used to monitor decisions of a neural network in operation at runtime.

SafeML would fall into this category of safety at runtime, as it measures the SDD during operation in comparison to the ECDF of datapoints seen at development time during training.
The proposed extension of SafeML, the Scope Compliance Uncertainty Estimators are meant as a method to evaluate the uncertainty of the operation design domain. Aside from a confidence measure in the model, it also provides a more situation aware scope compliance model than presented in \cite{klas2019uncertainty}.

\section{Methodology}
\label{sec:methodology}
In this section, we present our refined approach to address the limitations of SafeML through the introduction of a bootstrapped power analysis technique, the Epps and Singleton test and the uncertainty estimators. 
To experimentally validate our approach, we applied SafeML against an existing dataset, the GTSRB, and a CNN model trained for it.

\subsection{Bootstrapped Power Analysis}

\begin{figure}[t]
\centering
\includegraphics[width=0.9\textwidth]{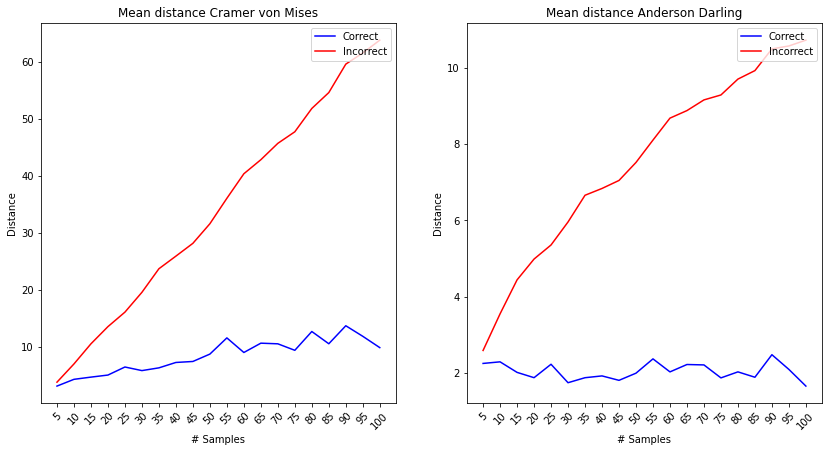}
\caption{SDD Over Varying Sampling Sizes for GTSRB} \label{fig:GTDSsmplVariance}
\end{figure}

In \cite{afarhad2022safeml}, the sample size dependency between the measured SDD and the accuracy of the model was investigated. It was found that the overall difference between the SDD of a batch of randomly sampled correctly classified points and incorrectly classified points increases, as the number of drawn samples increases, as shown in \cref{fig:GTDSsmplVariance}.

\begin{figure}[t]
\centering
\includegraphics[width=0.9\textwidth]{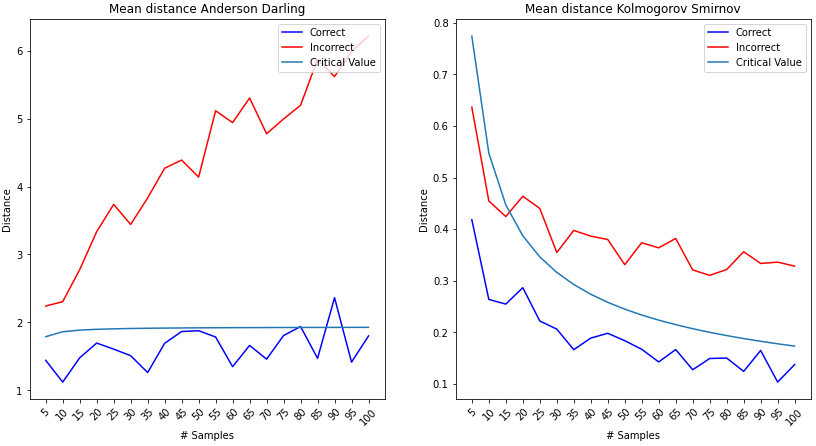}
\caption{SDD Over Varying Sampling Sizes with Corresponding Critical Value for GTSRB} \label{fig:critDiffSample}
\end{figure}

This property of distance measures can be exploited by overlaying the critical values of the corresponding statistical distance measure, as shown in \cref{fig:critDiffSample}, such that the point at which all the correct samples have been accepted, and all the incorrect samples have been rejected at the specified confidence level $\alpha$, is the point where a sufficient number of samples has been achieved. The obvious drawback is that the SDD was measured for randomly sampled points and might not reflect the true behaviour of the set.
By bootstrapping the above method for a number of trials, the drawback can be mitigated, such that the percentage of passed trials would be a reflection of the power of the test in question at that number of samples. According to Bonferroni \cite{bonferroni}, the chosen confidence level has to be adjusted depending on the number of trials, such that $1 - \frac{\alpha}{m}$ is used instead, where $m$ is the number of trials.

The critical values can only be easily calculated for Kolmogorov-Smirnov and Anderson-Darling for a high number of samples and a varying confidence level $\alpha$ \cite{crit_value_ks1976,pettitt_two-sample_1976}. Therefore, the bootstrap power analysis was only applied to them. This approach can be used for the other statistical distance measures by employing the aforementioned Monte Carlo simulations \cite{hall_bootstrap_1996} to obtain the critical value at the cost of high computational time.

This approach was validated by applying it to the synthetic dataset, generated with known parameters in \cite{afarhad2022safeml}, of an autonomous vehicle driving with a specific behavior profile. The t-test was used as a power analysis method returning 120 as the number of samples needed for a trustworthy prediction. However, using the proposed bootstrap power analysis approach, the number obtained was 140. The increase can be attributed to the non-parametric nature of the approach, as it can only return an estimate, as opposed to a concrete number. But this result can be considered as a validation of the method.


\subsection{ECF-based SSD}
The characteristics function is a Fourier-series transformation of the cumulative distribution function \cite{sebas2009}. 
\begin{equation}
    \label{eq:cf_conti}
    \phi_k(t) = E[e^{itx}] =  \int^{\infty}_{-\infty}e^{itx}dF_{n_k}
\end{equation}
Where $t$ is a real number, or series of numbers and $i = \sqrt{-1}$.

To make use of the characteristic function as a goodness of fit measure, the Epps and Sigleton test~(ES test) was proposed in \cite{es1986}. The transformation requires a set of parameters $t_1, t_2, . . . , t_j$, which should be calibrated to provide the test with sufficient power \cite{sebas2009}. However, simulations with nine different families of distributions were studied in \cite{es1986}, where using the set $t_1 = 0.4$ and $t_2 = 0.8$ with $(J = 2)$  was found to perform optimally. The implementation follows the approach described in \cite{sebas2009}, with the recommended factor $t_J = (0.4, 0.8)$.

The ES test statistic $W2$ is distributed asymptotically as $\chi^2$ \cite{sebas2009}. 
Through the known distribution of $\chi^2$ the p-value can be easily calculated without using a similar bootstrap approach for the other ECDF distance measures.

\begin{figure}[t]
\centering
\includegraphics[width=0.7\textwidth]{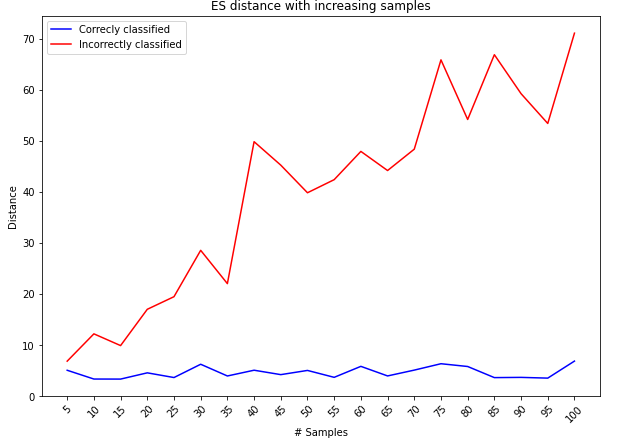}
\caption{SDD Over Varying Sampling Sizes for GTSRB - ES test} \label{fig:essampvar}
\end{figure}

%
Lastly, one of the significant properties of the ECDF method is the sample size dependency mentioned in \cref{sec:runsafe}, where the SDD difference between two distributions increases as the observed samples increase. The same experiment was conducted for the ES test, which was shown to follow the same trend as shown in \cref{fig:essampvar}. 
All of this makes the ES test a strong candidate to be used alongside the ECDF distance measures, or replace them in cases where it might outperform them. 

\subsection{SCUE}
\label{subsec:scue}
The current framework on which SafeML operates estimates a hard threshold upon which the trustworthiness of a given model's prediction is judged. By improving the SafeML metric into a continuous value, the acceptance threshold could be made conditional on the specific environment (for instance, making uncertainty for a pedestrian detector model to be more sensitive in school-zone areas), allowing it to be used as a situation-aware reliability measure. Moreover, it could be integrated into more holistic uncertainty measures, such as uncertainty wrappers \cite{klas2019uncertainty}, where a continuous scope compliance factor is needed.


By using a calibration set $\mathcal{C} \subset \mathbb{R} ^{m\times n}$, the SDD of samples in this calibration set could be correlated to the inaccuracy of the samples. Hence, under the definition that the uncertainty of a sample is the expected inaccuracy, building a function that would return the uncertainty from the measured SDD at runtime becomes a matter of finding a correlation between the measured SDD and the observed inaccuracy in the calibration set $\mathcal{C}$.

Constructing the calibration set requires fulfilling the requirement of covering points within the intended operational design domain~(ODD) and beyond it. Simply using misclassified points would not necessarily lead to a good estimate, especially given that their intended purpose is identifying previously unseen points. To this end, two methods to inject out of distribution~(OOD) samples have been identified:
\begin{itemize}
    \item \textit{Sample Corruption:} For many datatypes, there exist numerous corruption methods that could impact the model performance. This is especially relevant for image datasets in autonomous vehicle(AV) applications, as the vehicle could experience perception issues due to changes in weather, environment, .. etc.
    This method was used in building the SCUE for the GTSRB dataset, where a model was trained on the original images, and the estimators where calibrated and evaluated on corrupted images using image corruption methods introduced in \cite{corruption}.

    \item \textit{Class Exclusion:} In cases where a corruption is not possible, or not meaningful, a certain class of datatypes would be excluded such that the model would be trained on a part of the data, then the estimators would be calibrated and evaluated using the entire set.
\end{itemize}

The calibration set is then built by creating batches with increasingly higher numbers of OOD samples, such that it spans the entire range of accuracies uniformly, to follow the form:
    \begin{equation}
        \label{eq:design}
        \mathcal{C} = \bigg\{\mathbf{c}_i: \vert \mathbf{c}_{t,i}\vert = \frac{i}{m}n, \vert \mathbf{c}_{f,i} \vert = \frac{m - i}{m}n\bigg\}
    \end{equation}
where $\mathbf{c}_{t,i}$, and $\mathbf{c}_{f,i}$ are the subset of ODD-compliant and non-compliant samples in the $i$th batch of $\mathcal{C}$ respectively, $m$ is the number of batches in $\mathcal{C}$, and $n$ is the number of samples needed of a trustworthy measurement of SDD.\\
This can be achieved by using \cref{alg:buildset}, where the ratio of ODD-compliant samples to non-compliant samples is increased with the index of the current batch. \\
The algorithm follows the same notation as \cref{eq:design}. After concatenating the compliant and non-compliant subsets $\mathbf{c}_{t,i}$, and $\mathbf{c}_{f,i}$ into the batch $\mathbf{c}_{i}$, the batch is then shuffled to randomize the order in which the individual samples are observed. This ensures that the SDD would not be measured only on batches that start with $r = i/m$ percentage of compliant samples, but they are instead distributed in the batch.

\begin{algorithm}{\MYCALL{BuildSet}{$X_{cal}$, $\hat{y}$, $n$, $m$}}
\caption[Calibration Set Building Algorithm]{}
\label{alg:buildset}
\begin{algorithmic}[1]

\STATE $\mathcal{C} \leftarrow \MYNIL$ \algorithmiccomment{Set of m batches with n samples each}
\STATE $\mathcal{A} \leftarrow \MYNIL$\algorithmiccomment{Set of accuracies}

\LOOP
	\FOR{$i \leftarrow 0$ to  $m $}
        \STATE $r \leftarrow i / m$ \algorithmiccomment{Ratio of correct samples in batch i}
        \STATE $\mathbf{c}_{t,i} \leftarrow$ \MYCALL{Sample}{$X_{cal}[\hat{y} = 1]$, \MYCALL{Ceil}{$r*n$}}
        \STATE $\mathbf{c}_{f,i} \leftarrow$ \MYCALL{Sample}{$X_{cal}[\hat{y} = 0]$, \MYCALL{Ceil}{$(1 - r)*n$}}
        \STATE $\mathcal{C} \leftarrow  \mathcal{C} \cup$ \MYCALL{Shuffle}{$\mathbf{c}_{t,i} \cup \mathbf{c}_{f,i}$}
        \STATE $\mathcal{A} \leftarrow \mathcal{A} \cup \{r\}$
    \ENDFOR
\ENDLOOP

\RETURN $(\mathcal{C}, \mathcal{A})$
\end{algorithmic}
\end{algorithm}

The condition that the estimator function $\varphi(ssd)$ should fulfill is a monotonic increase in the output uncertainty as the measured SDD increases.
The correlation between the accuracy of the underlying model and the SDD measured was first observed in \cite{Aslansefat2020safeml}, later solidified in \cite{afarhad2022safeml}. If the coverage of the full range of accuracies in the calibration set is ensured, this property can be exploited, such that building the estimator becomes only a matter of fitting the correlation between the measured SDD and the observed inaccuracy in the calibration set $\varphi(ssd) = C(ssd, 1 - E[acc])$. Using the algorithm \cref{alg:buildset} with an increasing number of OOD samples ensures the accuracy coverage needed.

\begin{figure}[ht]
     \centering
     \begin{subfigure}[b]{0.48\textwidth}
         \centering
         \includegraphics[width=\textwidth]{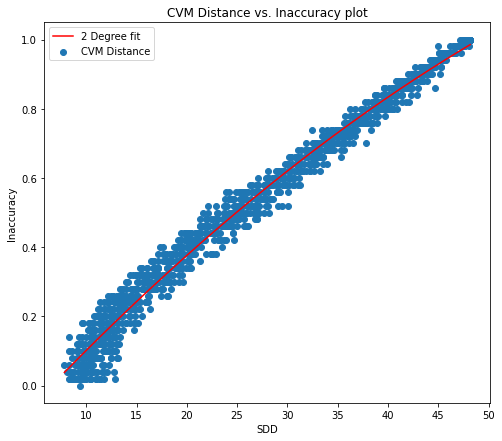}
         \caption{CVM Distance}
     \end{subfigure}
     \hfill
     \begin{subfigure}[b]{0.48\textwidth}
         \centering
         \includegraphics[width=\textwidth]{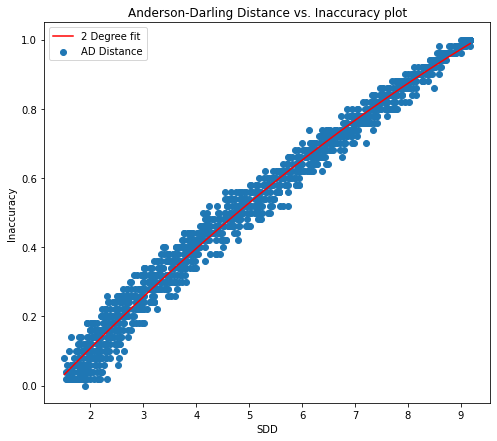}
         \caption{AD Distance}
     \end{subfigure}

     \begin{subfigure}[b]{0.48\textwidth}
         \centering
         \includegraphics[width=\textwidth]{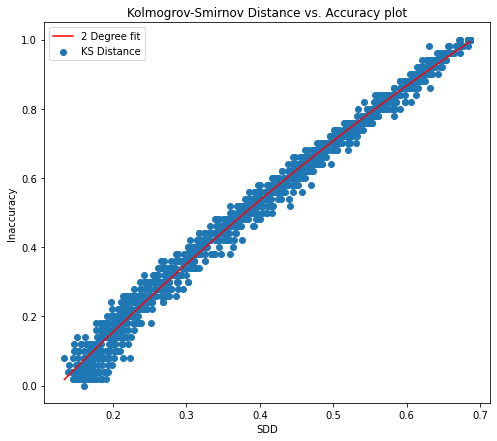}
         \caption{KS Distance}
     \end{subfigure}
     \hfill
     \begin{subfigure}[b]{0.48\textwidth}
         \centering
         \includegraphics[width=\textwidth]{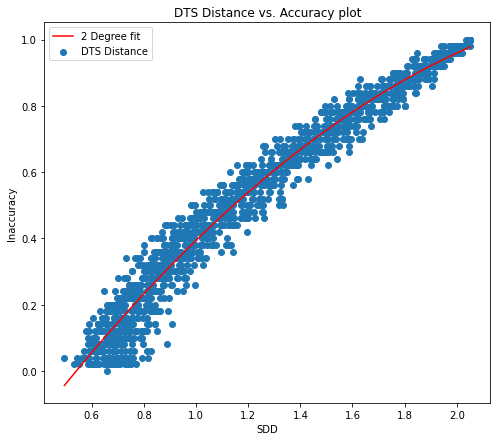}
         \caption{DTS Distance}
     \end{subfigure}

        \caption{Fit of the ECDF-based SDD estimators}
        \label{fig:ecdfEst}
\end{figure}

The ECDF-based estimators are all consistent in that the desired behaviour of correlation between the measured SDD and the observed inaccuracy is identical. As shown in \cref{fig:ecdfEst}, a polynomial of degree 2 is sufficient to capture the correlation. A test using the root mean square error~(RMSE) as metric was conducted to find the optimal choice of degree, finding that the decrease in RMSE does not justify the increase in complexity. 
All of the fitted functions are also bounded from above and below to the values observed during development time, such that any value below the minimum SDD measured at development time will result in $0$ uncertainty, and values above the max results in $1$ uncertainty.

\begin{equation}
    f(x)= 
\begin{cases}
    1,& \text{if } x > Max(sdd)\\
    0,  & \text{if } x > Min(sdd) \\
    SCUE(x), & \text{otherwise}
\end{cases}
\end{equation}

\begin{figure}[t]
     \centering
     \begin{subfigure}[b]{0.48\textwidth}
         \centering
         \includegraphics[width=\textwidth]{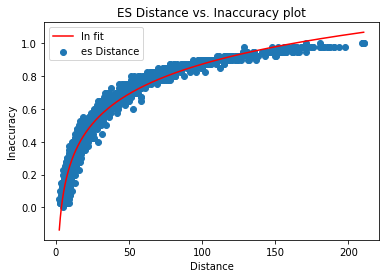}
         \caption{Log Fit}
     \end{subfigure}
     \hfill
     \begin{subfigure}[b]{0.48\textwidth}
         \centering
         \includegraphics[width=\textwidth]{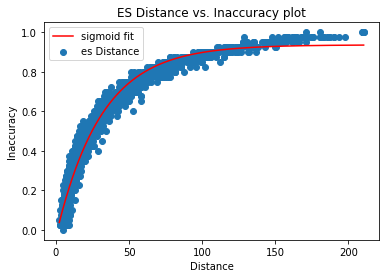}
         \caption{Sigmoid Fit}
     \end{subfigure}

        \caption{Fit of the ECDF-based SDD estimators}
        \label{fig:esEst}
\end{figure}

On the other hand, when the same approach is applied to the SDDs obtained from the ES test, a polynomial fit is no longer sufficient to capture the behaviour of the correlation. As shown in \cref{fig:esEst}, the correlation could be captured by using a fit of a logarithmic function, or a sigmoidal function. The RMSE and $R^2$ have been measured for both fits, where the RMSE is $0.3313, 0.3399$ for the logarithmic and sigmoidnal function respectively, and the $R^2$ is $0.8081, 0.6221$. 

Although no single fit is objectively the best, the logarithmic fit is chosen as it tends to capture a more conservative behaviour to the uncertainty, in that it is more taxing than the sigmoidal fit. The logarithm function fitted uses $3$ parameters according to the equation \cref{eq:log}. The function is bounded from above and below similarly to the other ECDF-based distance measures. However what differentiates the logarithm from the ECDF-based distance measures is, that is bounded from both above (maximum value of SDD returning $1$), and below (least defined value by the logarithm).

\begin{equation}
    \label{eq:log}
    f(x) = a.\log (x + b) + c
\end{equation}

\section{Experimental Evaluation}
\label{sec:experiment}
The evaluation of the method proposed is described out in this chapter. The settings and configurations are first presented, followed by a description of the method of evaluation. Finally, the results of the experiments are tabulated and discussed.

\subsection{Experimental Setup}
\label{subsec:expSetup}

The effectiveness of the Bootstrap Power analysis is tested for both the Kolmogorov-Smirnov distance and Anderson-Darling distance, as well as the newly introduced ES-test. 
An evaluation of the Scope Compliance Uncertainty Estimators~(SCUE) then tests their performance generally, as well as in comparison to the previous framework.

The proper evaluation of the method requires a number of steps, starting by building and training an ML model using the original dataset. The Calibration set is then constructed and estimators built according to \cref{subsec:scue}. Finally, the SDDs are calculated for the evaluation set along with their accuracies where a cost function is applied.

\textbf{Model Development:} Since the GTSRB is a set with relatively small image sizes, a custom CNN model is sufficient for the needed performance. The 12-layer architecture manages to achieve a 99.6\% validation accuracy. However, due to the corrupted test set, the accuracy of the test set is only $64\%$ for iteration 1, $63\%$ for iteration 2. 

\begin{figure}[ht]
\centering
\includegraphics[width=0.7\textwidth]{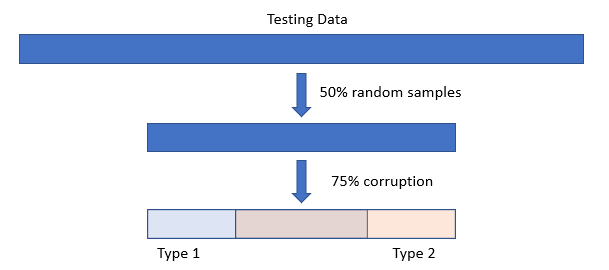}
\caption{Corruption Type Merging} \label{fig:corruptionMethod}

\end{figure}

\textbf{Calibration Set Construction:} In this experiment, the GTSRB is mainly used, where the OOD inclusion method used is sample corruption. To ensure that the estimators are not dependant on a certain type of corruption, a combination of 2 types is used as shown in \cref{fig:corruptionMethod}.

Half of the testing set is chosen randomly for corruption, where 75\% of it is corrupted with type 1, and another 75\% is corrupted with type 2, making an overlap of 50\% of samples that are corrupted with both types. In the experiment, the corruption of Fog and Frost are applied to the calibration set.

Using a completely independent set of images, the same corruption method is applied, and the sliding window approach is used to construct a set of the same shape as the calibration set. The SDD of each element of this set is measured and applied to the SCUE to retrieve its corresponding uncertainty. By setting a threshold for uncertainty, batches are rejected based on their uncertainty. This approach was used on the independent evaluation set with 3 different iterations as follows:
\begin{itemize}
    \item Calibration Context~(CC): the corruption types are the same as calibration (Frost - Fog)
    
    \item Semi-Out of Context~(SOC): only one corruption is the same as calibration (Frost - Snow)
    
    \item Completely Out of Context~(COC): the corruption types are completely different than calibration (Motion blur-Defocus blur)
\end{itemize}

\textbf{SDD Calculation:} Since the distance measures are univariate, the SDD must be calculated as a set of features. Using the individual pixels as features can lead to poor results, as the individual pixel would not be representative features of the image. Therefore, the output of the last layer of the CNN is used as the set of features instead. While it could be argued that this method makes this approach model-dependant
, the CNN is only used as a feature extractor and allows the exploitation of convolutional layers without needing to train an auxiliary Autoencoder network to extract the same features. 

\begin{figure}[ht]
     \centering

        \includegraphics[width=\textwidth]{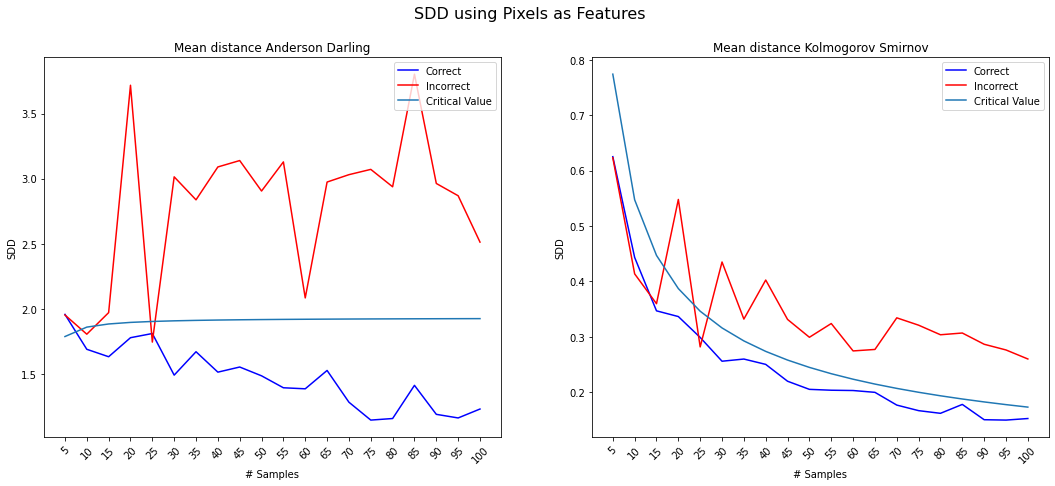}

        \caption{SDD from Pixels Features}
        \label{fig:pixCNN}
\end{figure}
\begin{figure}[ht]
     \centering

        \includegraphics[width=\textwidth]{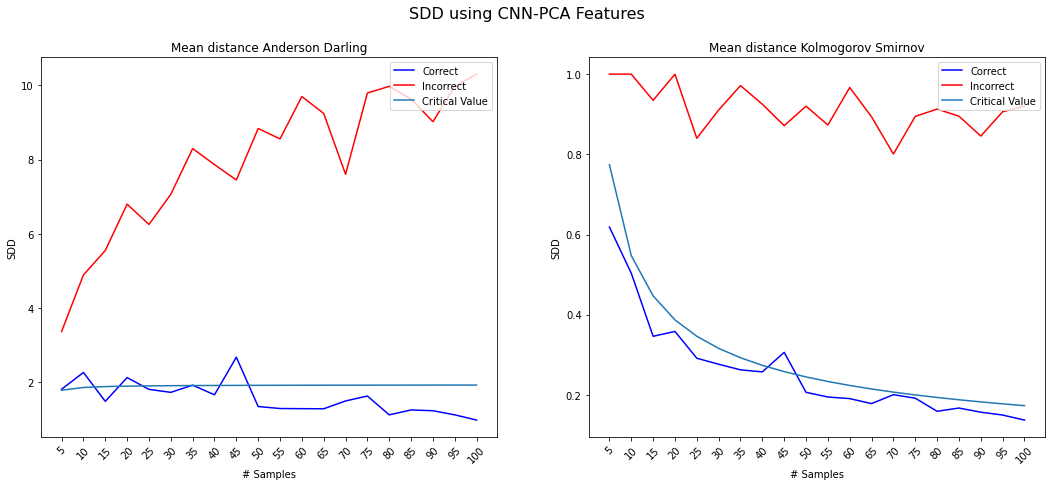}
        \caption{SDD from CNN-PCA Features}
        \label{fig:CNNfeat}
\end{figure}

Furthermore, to increase the computation speed, Principal Component Analysis~(PCA) is applied to the 100 features of the last layer and a number of components is taken to explain at least 85\% of the variance, which is 11 features. In \cref{fig:pixCNN}, the SDD calculated from the individual pixels is shown. While there is a small distinction between the 2 distances and using the individual pixels could lead to a separation of the OOD data, filtering the image through the CNN and reducing the dimensionality with PCA not only reduces the speed of computation, but increases the distinction as well, as shown in \cref{fig:CNNfeat}.

\subsection{Results}
\label{subsec:results}

Multiple iterations of the corrupted evaluation set were conducted and but only some of their results are shown and discussed in this section. Further examples are continued in the GitLab instance provided.\\

\begin{figure}[h]
\centering
\includegraphics[width=0.9\textwidth]{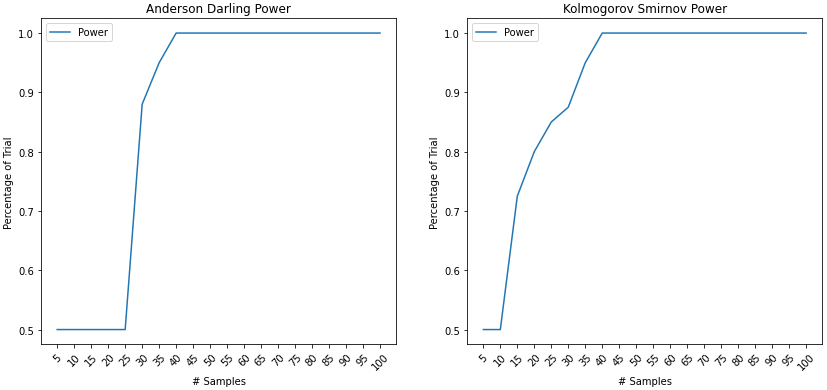}
\caption{Power of KS and AD Test} \label{fig:BPAKA}

\end{figure}

\textbf{Bootstrap Power Analysis Results:}
Using the Bootstrap power analysis method, the number of needed samples for a trustworthy SDD measurement can be found. By setting the confidence level $\alpha = 0.1$, the critical values for both Kolmogorov-Smirnov and Anderson-Darling can be calculated and compared to the measured SDD. The bootstrapping resulted in the power level shown in \cref{fig:BPAKA}.

\begin{figure}[h]
\centering
\includegraphics[width=0.55\textwidth]{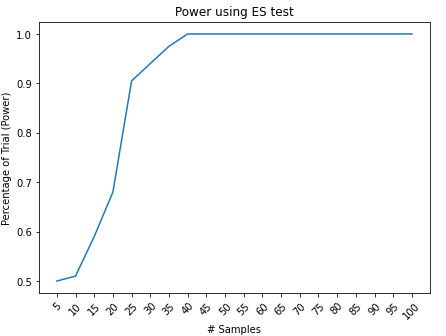}
\caption{Power of ES Test} \label{fig:BPAKS}

\end{figure}

The confidence level $\alpha=0.1$ was chosen as it is a common standard for statistical power levels. 
As for the ES test, the p-value can be directly used and compared to the confidence level, returning the power level shown in \cref{fig:BPAKS}. Similar to the other distance measures, the returned number where the power becomes 100\% is also the same.

The choice of number between the three measures, if they do not agree, becomes a matter of design; such that designing for a worst-case scenario means choosing the highest of the numbers. Otherwise, if a certain statistical distance measure were to be chosen as the focus, that would be the choice. For the choice of samples for the GTSRB, 50 samples were chosen for the evaluation.\\

\begin{figure}[ht]

\centering
\includegraphics[width=\textwidth]{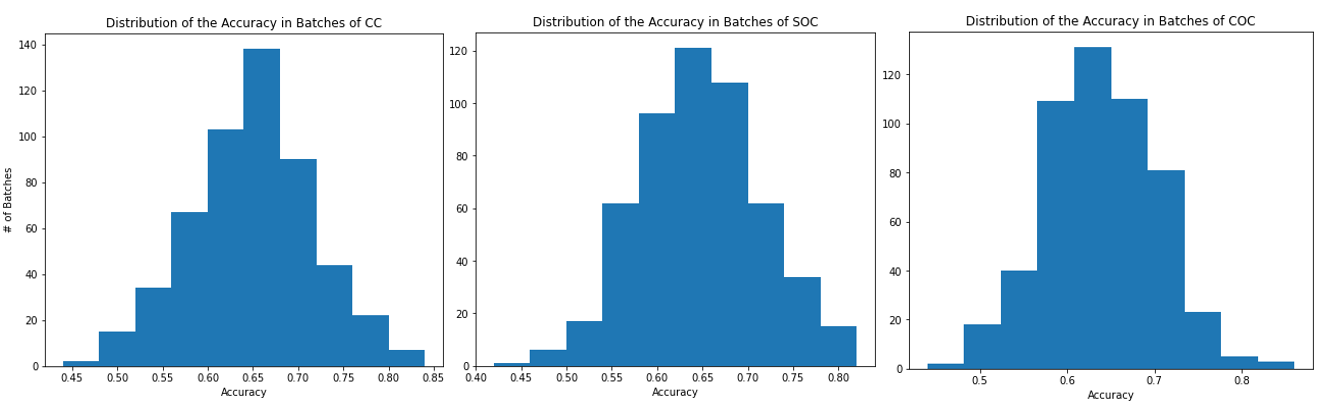}
\caption{Accuracy of Batches in Evaluation Each Set} \label{fig:iter1Acc}

\end{figure}

\textbf{SCUE Evaluation}
The result of evaluating the fitted estimators for the GTSRB follows. The distribution of accuracies for each of the datasets is shown in \cref{fig:iter1Acc}. While the distribution shape varies between set to set, the overall range is quite similar.

By measuring the SDD of those sets and applying the estimators, with an uncertainty threshold of $0.5$, which is chosen arbitrarily, the result in \cref{tab:rejFF6} is obtained. The number of points rejected by each statistical distance measure is shown, and how many points of which were falsely rejected (their inaccuracy is lower than $0.5$) and how many batches with an inaccuracy higher than $0.5$ are missed, and deemed as not too distant.

\begin{table}[h]
\caption{SCUE Confusion Matrix with Different Datasets}
\begin{center}

\begin{tabular*}{\textwidth}{@{\extracolsep{\fill}}>{\scriptsize}l|>{\scriptsize}c>{\scriptsize}c>{\scriptsize}c|>{\scriptsize}c>{\scriptsize}c>{\scriptsize}c|>{\scriptsize}c>{\scriptsize}c>{\scriptsize}c} 
\hline
& \multicolumn{9}{>{\scriptsize}c}{Dataset} \\\hline
     & \multicolumn{3}{>{\scriptsize}c}{CC} & \multicolumn{3}{>{\scriptsize}c}{SOC} &                      \multicolumn{3}{>{\scriptsize}c}{COC}  \\\hline
     & \bfseries Rej  &  \bfseries FR &  \bfseries Missed  &  \bfseries Rej  & \bfseries FR & \bfseries Missed  &  \bfseries Rej  &  \bfseries FR & \bfseries Missed\\\hline\hline
{ \bfseries CVM}      & 9 & 3 & 2     & 12 & 6 & 2  & 15 & 7 & 0 \\\hline
{ \bfseries AD}     & 10 & 4 & 2     & 16 & 10 & 2  & 15 & 7 & 0  \\\hline
{ \bfseries KS}        & 11 & 0 & 1     & 14 & 7 & 3  & 13 & 5 & 0  \\\hline
{ \bfseries WS}        & 14 & 7 & 2     & 20 & 13 & 1  & 16 & 8 & 0  \\\hline
{ \bfseries DTS}        & 17 & 9 & 1     & 19 & 14 & 3  & 23 & 17 & 1  \\\hline
{ \bfseries ES}       & 11 & 4 & 1     & 10 & 6 & 4  & 16 & 9 & 0  \\\hline

\end{tabular*}

\label{tab:rejFF6}
\end{center}
\end{table}



Finally, the performance of the SCUEs is compared to the SafeML threshold method. The values published originally in \cite{afarhad2022safeml} are used. However, these thresholds were calculated using the class exclusion method. Therefore, new thresholds were calculated and fine-tuned (under the assumption of $0.6$ is the boundary of acceptable points). To make an equivalent comparison, the values use an uncertainty threshold of $0.6$, taking the same assumption in calculating the SafeML threshold. The calibration context set was used in \cref{tab:safeCC}.

\begin{figure}[ht]
\centering
\includegraphics[width=0.67\textwidth]{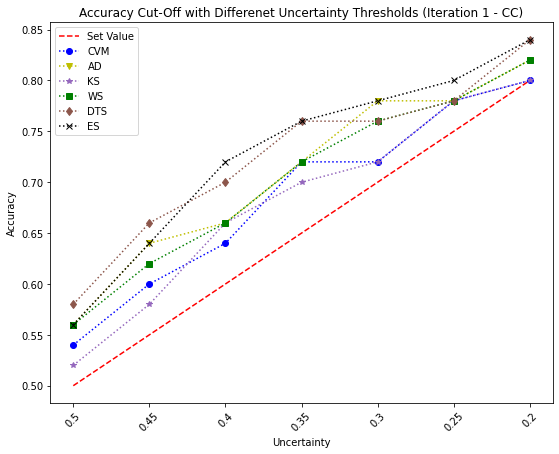}
\caption{Uncertainty Threshold against Accuracy Cut-Off (CC)} \label{fig:accCutCC}

\end{figure}

\begin{table}[h]
\caption{SCUE Comparison to SafeML threshold Method using CC set}
\begin{center}

\begin{tabular*}{\textwidth}{@{\extracolsep{\fill}}>{\scriptsize}l|>{\scriptsize}c>{\scriptsize}c>{\scriptsize}c|>{\scriptsize}c>{\scriptsize}c>{\scriptsize}c|>{\scriptsize}c>{\scriptsize}c>{\scriptsize}c|>{\scriptsize}c>{\scriptsize}c>{\scriptsize}c} 
\hline
& \multicolumn{12}{>{\scriptsize}c}{Dataset} \\\hline
    & \multicolumn{3}{>{\scriptsize}c}{SCUE - 0.5} & \multicolumn{3}{>{\scriptsize}c}{SCUE - 0.4} & \multicolumn{3}{>{\scriptsize}c}{SafeML} &  \multicolumn{3}{>{\scriptsize}c}{SafeML - FT}  \\\hline
     & \bfseries Rej  &  \bfseries FR &  \bfseries Missed & \bfseries Rej  &  \bfseries FR &  \bfseries Missed  &  \bfseries Rej  & \bfseries FR & \bfseries Missed  &  \bfseries Rej  &  \bfseries FR & \bfseries Missed\\\hline\hline
{ \bfseries CVM}    & 9 & 3 & 2     & 125 & 16 & 23    & 0 & 0 & 104    & 3 & 1 & 103 \\\hline
{ \bfseries AD}     & 10 & 4 & 2     & 133 & 19 & 21    & 61 & 53 & 102    & 7 & 5 & 103\\\hline
{ \bfseries KS}     & 11 & 0 & 1     & 139 & 16 & 15    & 89 & 72 & 97    & 0 & 0 & 104 \\\hline
{ \bfseries WS}     & 14 & 7 & 2     & 129 & 20 & 27    & 517 & 353  & 0    & 71 & 61 & 100\\\hline
{ \bfseries DTS}    & 17 & 9 & 1     & 137 & 27 & 29    & - & - & -    & 42 & 37 & 102\\\hline
{ \bfseries ES}     & 11 & 4 & 1     & 154 & 35 & 23    & - & - & -     & - & - & -\\\hline

\end{tabular*}

\label{tab:safeCC}
\end{center}
\end{table}
Since the SCUEs seem to reject more samples than required in most cases, their expected performance against certain thresholds is plotted in \cref{fig:accCutCC}, where the accuracy cut-off point is shown against the chosen uncertainty i.e. when the chosen uncertainty threshold is $0.4$, the cut-off should be $0.6$ and so on.

\section{Discussion}
While there is no formal proof to show the validity of the results obtained from the bootstrap power analysis technique proposed here, the fact that all of the three statistical distance measures, Kolmogorov-Smirnov, Anderson-Darling, and Epps-Singleton agree on the same result, is a good indication that the number obtained is not arbitrary. Furthermore, when the bootstrap power analysis method was applied to a dataset with known parameters, the result corroborated with that obtained from using the parametric approach, t-test. From all this, it could be concluded that the bootstrap power analysis technique is a suitable non-parametric alternative to other power analysis tests.

Even though using the ECF directly did not yield the desired effect of reducing the size of the data needed to measure the SDD, it is still beneficial, as it can be used through the ES test. The ES test fulfills all the properties of the ECDF-based statistical distance measures and comes with the added advantage of following the $\chi^2$ distribution. Through this distribution, the p-value can be calculated without the need to spend computational resources to bootstrap and also allows it to be used in the bootstrap power analysis.

The performance of the SCUEs was evaluated on multiple variations of the GTSRB with different added levels of OOD of the applied image corruptions. However, the KS-based uncertainty estimator was constantly performing the most conservatively amongst the other distance measures. This can also be noticed in \cref{fig:ecdfEst}, where the KS fit is the tightest. This could be attributed to its relatively simple formula being only the supremum value. The CVM-based estimator also shares a similar behavior. On the other hand, DTS and Wasserstein \cite{dowd_new_2020} were the most aggressive, rejecting the most batches. However, it should be noted that this observation is not generalizable to all use-cases, as the performance of the statistical distances is highly dependant on the data and its distribution. While the conservative behavior of the KS- and CVM-based estimators might not be the generally desired behavior, it could be more impactful in situations were a false negative is much more costly and poignant than a false positive (such as attacks, or fraud detection).  

Furthermore, the logarithmic fit of the ES-based uncertainty estimator would seem, on first insight, to reflect a more accurate behavior of uncertainty. 
But during evaluation, the ES-based uncertainty estimator was not amongst the top performing estimators. This can be attributed to the limited range of accuracies present in the evaluation set.
On its own, the SCUE method did prove to be effective in detecting OOD points. However, when compared to the previous framework, its advantages become much more prominent as it addresses most of the issues hindering the previous approach.

\section{Conclusion and Future Work}
\label{sec:conclusion}
In this paper, the established framework for using the statistical distance measures in SafeML was investigated and its limitations were identified. In an attempt to improve that framework, the contribution of this work is three-fold. 

\begin{enumerate}
    \item Removed the inherent distribution assumption in the previous framework and ensured a completely non-parametric approach. 
    \item Increased the speed of the computation of the SDD by investigating the use of the ECF instead of the ECDF, through the ES test.
    \item Developed a seamless approach to establish trust in the observed datasets in operation, by removing the necessity of making various design decisions and creating a bounded continuous output in the form of a SCUE, that can be integrated into more holistic approaches.
\end{enumerate}

The methods and approaches were developed using 2 different datasets, the GTSRB, which is a collection of images used for classification.
Experiments were conducted to test the performance of the SCUE to evaluate its effectiveness even when new OOD inputs are encountered.

While the experiments in this thesis showed that a consistent performance behaviour pattern was apparent, it does not necessarily imply that this is generalizable to other kinds of datasets and problems. The SCUEs would benefit from being tested with different datasets with different combinations of approaches of OOD injection.

Furthermore, both datasets used have a clear definition of ground-truth and a false prediction, which allowed straightforward construction of a calibration set as introduced in \cref{subsec:scue}. However, in problems such as segmentation, object recognition, .. etc., a clear binary true and false prediction is not available. The method must be then adapted to the problem. This also partly applies to datasets with chronological context, where a sliding window is the only method available to build the calibration set and it cannot ensure a set that covers the entire span of accuracies needed.

Since all of the statistical distance measures, both ECDF-based and ECF, are univariate, the SDD for a sample with multiple features is calculated by taking the average of all the features. This can be mitigated by investigating the multivariate version of the distance measures, or by adding another layer to the calculation, such as conformal prediction, as the multivariate distance measures tend to be computationally complex. 


\section*{Code Availability}
Regarding the research reproducibility, codes and functions supporting this paper are published online at: \url{https://tinyurl.com/muw4jpc3} and \url{https://github.com/ISorokos/SafeML}.

%
%
\bibliographystyle{splncs04}

\begin{thebibliography}{10}
\providecommand{\url}[1]{\texttt{#1}}
\providecommand{\urlprefix}{URL }
\providecommand{\doi}[1]{https://doi.org/#1}

\bibitem{akram_stadre_2022}
Akram, M.N., Ambekar, A., Sorokos, I., Aslansefat, K., Schneider, D.: {StaDRe} and~{StaDRo}: {Reliability} and {Robustness} {Estimation} of {ML}-{Based} {Forecasting} {Using} {Statistical} {Distance} {Measures}. In: Computer {Safety}, {Reliability}, and {Security}. {SAFECOMP} 2022 {Workshops}. pp. 289--301 (2022)

\bibitem{aslansefat2021toward}
Aslansefat, K., Kabir, S., Abdullatif, A., Vasudevan, V., Papadopoulos, Y.: Toward improving confidence in autonomous vehicle software: A study on traffic sign recognition systems. Computer  \textbf{54}(8),  66--76 (2021)

\bibitem{Aslansefat2020safeml}
Aslansefat, K., Sorokos, I., Whiting, D., Tavakoli~Kolagari, R., Papadopoulos, Y.: Safeml: Safety monitoring of machine learning classifiers through statistical difference measures. In: Model-Based Safety and Assessment. pp. 197--211. Springer International Publishing (2020)

\bibitem{cheng2019nn}
Cheng, C.H., Huang, C.H., N{\"u}hrenberg, G.: nn-dependability-kit: Engineering neural networks for safety-critical autonomous driving systems. In: Int. Conf. on Computer-Aided Design (ICCAD). pp.~1--6. IEEE (2019)

\bibitem{dowd_new_2020}
Dowd, C.: A {New} {ECDF} {Two}-{Sample} {Test} {Statistic}. Tech. rep. (Jul 2020), \url{http://arxiv.org/abs/2007.01360}

\bibitem{bonferroni}
Dunn, O.J.: Multiple comparisons among means. Journal of the American Statistical Association  \textbf{56}(293),  52--64 (1961), \url{http://www.jstor.org/stable/2282330}

\bibitem{es1986}
Epps, T., Singleton, K.J.: An omnibus test for the two-sample problem using the empirical characteristic function. Journal of Statistical Computation and Simulation  \textbf{26}(3-4),  177--203 (1986). \doi{10.1080/00949658608810963}

\bibitem{afarhad2022safeml}
Farhad, A.H., Sorokos, I., Schmidt, A., Akram, M.N., Aslansefat, K., Schneider, D.: Keep {Your} {Distance}: {Determining} {Sampling} and~{Distance} {Thresholds} in {Machine} {Learning} {Monitoring}. In: Model-{Based} {Safety} and {Assessment}. pp. 219--234. Springer International Publishing (2022)

\bibitem{fischer2019dl2}
Fischer, M., Balunovic, M., Drachsler-Cohen, D., Gehr, T., Zhang, C., Vechev, M.: Dl2: training and querying neural networks with logic. In: International Conference on Machine Learning. pp. 1931--1941. PMLR (2019)

\bibitem{crit_value_ks1976}
Gail, M.H., Green, S.B.: Critical values for the one-sided two-sample kolmogorov-smirnov statistic. Journal of the American Statistical Association  \textbf{71}(355),  757--760 (1976). \doi{10.1080/01621459.1976.10481562}

\bibitem{gerasimou2020importance}
Gerasimou, S., Eniser, H.F., Sen, A., Cakan, A.: Importance-driven deep learning system testing. In: 42nd Int. Conf. on Software Engineering (ICSE). IEEE (2020)

\bibitem{sebas2009}
Goerg, S.J., Kaiser, J.: Nonparametric testing of distributions—the epps–singleton two-sample test using the empirical characteristic function. The Stata Journal  \textbf{9}(3),  454--465 (2009). \doi{10.1177/1536867X0900900307}

\bibitem{Gros2020Modelchecking}
Gros, T.P., Hermanns, H., Hoffmann, J., Klauck, M., Steinmetz, M.: Deep statistical model checking. In: Gotsman, A., Sokolova, A. (eds.) Formal Techniques for Distributed Objects, Components, and Systems. pp. 96--114. Springer International Publishing, Cham (2020)

\bibitem{hall_bootstrap_1996}
Hall, P., Horowitz, J.L.: Bootstrap {Critical} {Values} for {Tests} {Based} on {Generalized}-{Method}-of-{Moments} {Estimators}. Econometrica  \textbf{64}(4),  891--916 (1996)

\bibitem{corruption}
Hendrycks, D., Dietterich, T.G.: Benchmarking neural network robustness to common corruptions and perturbations. ArXiv  \textbf{abs/1903.12261} (2018)

\bibitem{henne2020benchmarking}
Henne, M., Schwaiger, A., Roscher, K., Weiss, G.: Benchmarking uncertainty estimation methods for deep learning with safety-related metrics. In: SafeAI@ AAAI. pp. 83--90 (2020)

\bibitem{klas2020framework}
Kl{\"a}s, M., J{\"o}ckel, L.: A framework for building uncertainty wrappers for ai/ml-based data-driven components. In: International Conference on Computer Safety, Reliability, and Security. pp. 315--327. Springer (2020)

\bibitem{klas2019uncertainty}
Kl{\"a}s, M., Sembach, L.: Uncertainty wrappers for data-driven models. In: International Conference on Computer Safety, Reliability, and Security. pp. 358--364. Springer (2019)

\bibitem{mirman2019provable}
Mirman, M., Singh, G., Vechev, M.: A provable defense for deep residual networks. arXiv preprint arXiv:1903.12519  (2019)

\bibitem{paterson2021detection}
Paterson, C., Calinescu, R., Picardi, C.: Detection and mitigation of rare subclasses in deep neural network classifiers. In: 2021 IEEE International Conference On Artificial Intelligence Testing (AITest). pp. 9--16. IEEE Computer Society, Los Alamitos, CA, USA (aug 2021)

\bibitem{pettitt_two-sample_1976}
Pettit, A.N.: A two-sample {Anderson}-{Darling} rank statistic. Biometrika  \textbf{63}(1),  161--168 (1976). \doi{10.1093/biomet/63.1.161}

\bibitem{ruoss2020efficient}
Ruoss, A., Baader, M., Balunovi{\'c}, M., Vechev, M.: Efficient certification of spatial robustness. arXiv preprint arXiv:2009.09318  (2020)

\bibitem{shafaei_uncertainty_2018}
Shafaei, S., Kugele, S., Osman, M.H., Knoll, A.: Uncertainty in {Machine} {Learning}: {A} {Safety} {Perspective} on {Autonomous} {Driving}. In: Computer {Safety}, {Reliability}, and {Security}. pp. 458--464. Springer International Publishing, Cham (2018)

\bibitem{sharafaldin2018toward}
Sharafaldin, I., Lashkari, A.H., Ghorbani, A.A.: Toward generating a new intrusion detection dataset and intrusion traffic characterization. ICISSp  \textbf{1},  108--116 (2018)

\bibitem{singh2019certifyNN}
Singh, G., Gehr, T., Püschel, M., Vechev, M.: An abstract domain for certifying neural networks. Proceedings of the ACM on Programming Languages  \textbf{3},  1--30 (01 2019). \doi{10.1145/3290354}

\bibitem{stallkamp2012man}
Stallkamp, J., Schlipsing, M., Salmen, J., Igel, C.: Man vs. computer: Benchmarking machine learning algorithms for traffic sign recognition. Neural networks  \textbf{32},  323--332 (2012)

\bibitem{aiFailure}
Yampolskiy, R., Spellchecker, M.: Artificial intelligence safety and cybersecurity: a timeline of ai failures  (10 2016)

\end{thebibliography}
%

\end{document}